\documentclass[conference]{IEEEtran}
\IEEEoverridecommandlockouts
\usepackage{url}
\usepackage{booktabs}
\usepackage{multirow}
\usepackage{cite}
\usepackage{amsmath,amssymb,amsfonts}
\usepackage{algorithmic}
\usepackage{graphicx}
\usepackage{textcomp}
\usepackage{xcolor}
\usepackage{nicematrix}
\usepackage[a4paper, total={184mm,239mm}]{geometry}
\def\BibTeX{{\rm B\kern-.05em{\sc i\kern-.025em b}\kern-.08em
    T\kern-.1667em\lower.7ex\hbox{E}\kern-.125emX}}
\begin{document}

\title{LithoSeg: A Coarse-to-Fine Framework for High-Precision Lithography Segmentation}

\author{Xinyu He, Botong Zhao, Bingbing Li, Shujing Lyu, Jiwei Shen, Yue Lu}


\maketitle

\begin{abstract}
Accurate segmentation and measurement of lithography scanning electron microscope (SEM) images are crucial for ensuring precise process control, optimizing device performance, and advancing semiconductor manufacturing yield. Lithography segmentation requires pixel-level delineation of groove contours and consistent performance across diverse pattern geometries and process window. However, existing methods often lack the necessary precision and robustness, limiting their practical applicability. To overcome this challenge, we propose LithoSeg, a coarse-to-fine network tailored for lithography segmentation. In the coarse stage, we introduce a Human-in-the-Loop Bootstrapping scheme for the Segment Anything Model (SAM) to attain robustness with minimal supervision. In the subsequent fine stage, we recast 2D segmentation as 1D regression problem by sampling groove-normal profiles using the coarse mask and performing point-wise refinement with a lightweight MLP. LithoSeg outperforms previous approaches in both segmentation accuracy and metrology precision while requiring less supervision, offering promising prospects for real-world applications.
\end{abstract}


\section{Introduction}
The continuous drive for higher integrated circuit (IC) performance has pushed semiconductor manufacturing to ever-smaller feature sizes. At these advanced nodes, the complexity of lithography physics and circuit patterns makes meticulous control over lithography parameters and process windows~\cite{Bhattarai2017}. Therefore, lithography segmentation gained increasing attention for its ability to perform process control and optimization. Precise segmentation of lithography groove patterns enables the identification and quantification of lithography quality that directly impact device performance and yield~\cite{SEM_image_contouring_for_OPC, Pix2Pix-HeXinyu}.

Traditionally, lithography segmentation has depended on skilled technicians to visually inspect and perform measurements, a process that is both labor-intensive and susceptible to human mistakes \cite{Bunday2004}. On the other hand, the practical deployment of deep learning for lithography segmentation is hindered by two principal factors~\cite{UISS, ronneberger2015unetconvolutionalnetworksbiomedical, SegViT, Xu2025SegNetOPC, EUnetpp}. First, these models require extensive, densely annotated training data, the generation of which is prohibitively costly and time-consuming. This data dependency becomes particularly acute with the advancement of manufacturing nodes, necessitating repeated data collection and training efforts for each new technology. Second, the outputs from these networks often exhibit overly smoothed contours. This characteristic is a significant drawback, as it makes the models inadequate for assessing lithography quality, where precise measurement of contour roughness is crucial~\cite{Jacob2023, learning_to_detect_defect}.

To address these fundamental challenges, we propose LithoSeg, a robust, high-precision, coarse-to-fine lithography segmentation network with minimal annotation overhead. In the coarse stage, we leverage a vision foundation model, SAM~\cite{Kirillov_2023_ICCV_SAM}, and adapt it to the lithography domain through a Human-in-the-Loop Bootstrapping scheme. Specifically, we first extract bounding boxes of grooves from layout images and use them as prompts for SAM to generate initial coarse segmentation masks. These masks are then refined with human supervision by removing clearly erroneous outputs. The cleaned dataset is subsequently used to fine-tune SAM without prompts, and this process is iteratively repeated to progressively enhance model performance in the target domain. This bootstrapping strategy significantly reduces both annotation cost and training time, while the inherent generalization capability of vision foundation models ensures robustness to unseen patterns and process window.

\begin{figure*}[t!]
    \centering
    \includegraphics[width=0.95\linewidth]{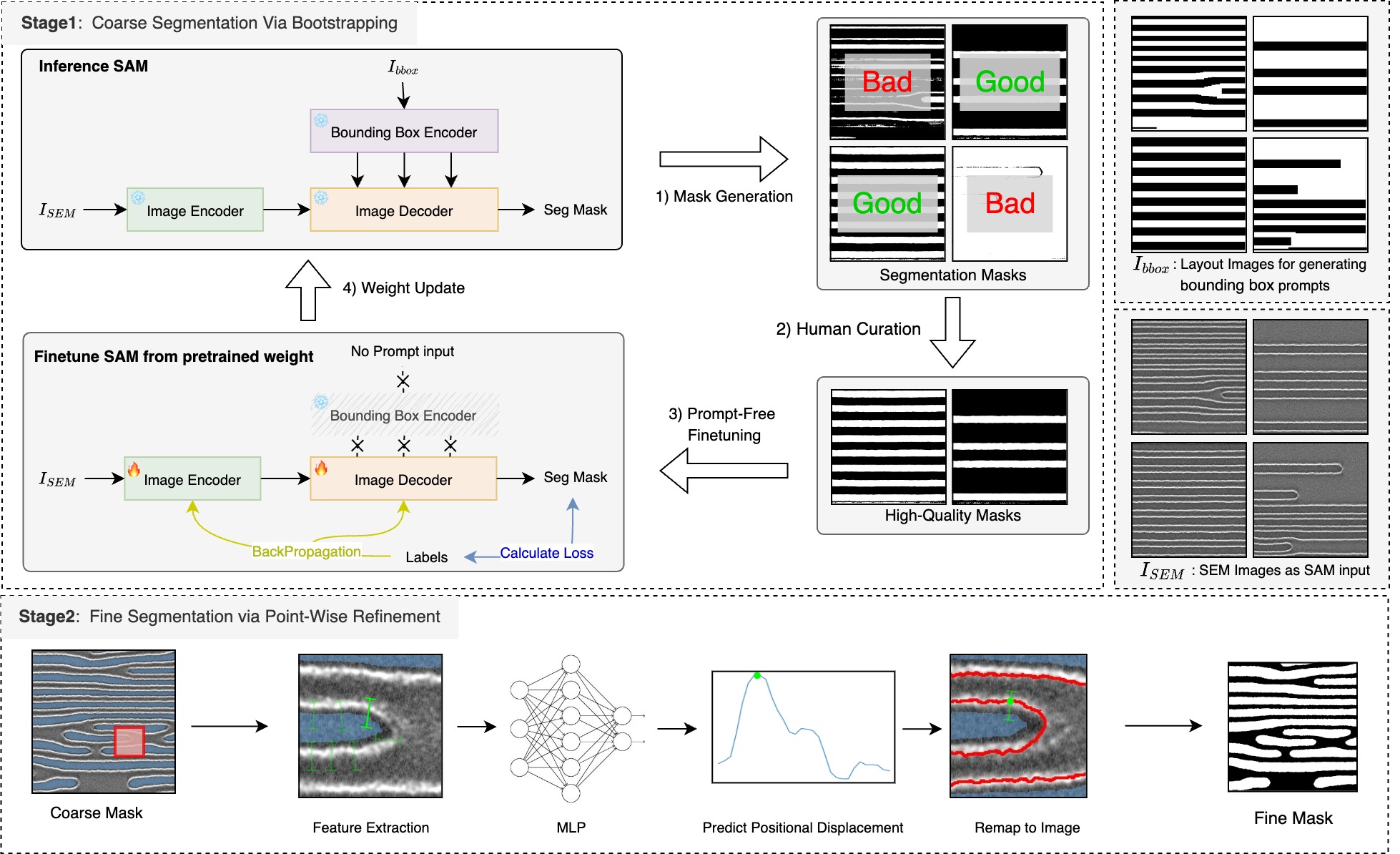}
    \caption{This figure illustrates the proposed LithoSeg framework: a two-step process for metrology-ready segmentation of SEM images. Stage 1 (Coarse Segmentation via human-in-the-loop Bootstrapping) involves using SAM to generate initial segmentation masks, followed by Human Curation to classify good and bad results, which are then used to finetune SAM from pretrained weight. Stage 2 (Fine-Grained Contour Segmentation) refines the coarse segmentation by processing each contour point, predicting contour fineness, and generating a detailed fine segmentation mask.}
    \label{fig:mainFig}
\end{figure*}

In the second stage, we address the precision challenge by reformulating the 2D segmentation problem as multiple point-wise 1D regression tasks. In particular, we extract 1D groove features along the normal directions of the coarse segmentation mask and feed them into a Multi-Layer Perceptron (MLP) network to predict the positional displacement of each contour point. This Point-Wise Refinement approach achieves pixel-level accuracy while maintaining strong generalization and incurring minimal additional computational overhead. Extensive experiments demonstrate that LithoSeg surpasses state-of-the-art methods in both segmentation and metrology accuracy, particularly when applied to novel patterns, new processes, and sub-optimal lithography parameters.

Our key contributions can be summarized as follows:
\begin{itemize}
\item We propose LithoSeg, a novel, generalizable coarse-to-fine network for lithography segmentation that balances generalization capability and metrology-level precision with minimal annotation overhead.
\item We propose a Human-in-the-Loop Bootstrapping strategy that leverages SAM to perform coarse segmentation with minimal supervision, significantly reducing annotation cost.
\item We propose a point-wise refinement strategy that reformulates 2D segmentation into 1D regression problem, enabling pixel-level accuracy with minimal computational overhead.
\end{itemize}

\section{Method}

\subsection{Framework Overview}

The primary goal of this work is to achieve accurate and robust lithography segmentation to support lithography metrology. We formulate this task as a binary segmentation problem: given a Scanning Electron Microscope (SEM) image \( I_{\text{SEM}} \in \mathbb{R}^{H \times W} \), the goal is to produce a binary segmentation mask \( M \in \{0,1\}^{H \times W} \) that precisely delineates the photoresist patterns.

As illustrated in Fig.~\ref{fig:mainFig}, the proposed LithoSeg framework follows a two-stage design. The first stage performs coarse segmentation by leveraging SAM within a Human-in-the-Loop Bootstrapping scheme, substantially reducing annotation cost. The second stage applies a Point-Wise Refinement strategy to correct contour displacements, thereby achieving pixel-level precision with minimal computational overhead.

\begin{figure*}
     \centering
     \includegraphics[width=\linewidth]{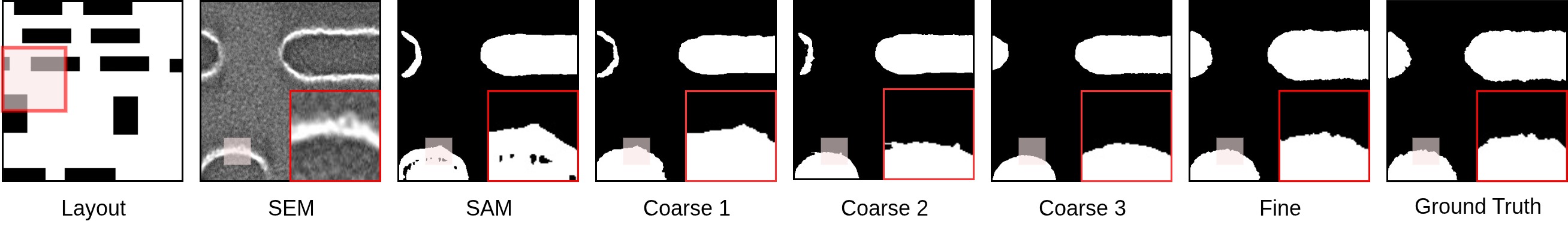}
     \vspace{-20pt} 
     \caption{Visualization of LithoSeg's input, output and intermediate outputs. This coarse-to-fine pipeline progressively reduces artifacts and sharpens boundaries, closely matching the ground truth mask and its boundary roughness. }
     \label{fig:visual_step}
 \end{figure*}

\subsection{Coarse Segmentation via Human-in-the-Loop Bootstrapping}

In the coarse segmentation stage, we employs an iterative bootstrapping strategy designed to progressively adapt SAM to the lithography segmentation domain with minimal human supervision. This stage comprises four steps:
\subsubsection{Mask Generation}: We start the process by utilizing layout images to generate bounding boxes $I_{bbox}$ for the photoresist trenches. SAM uses those bounding boxes to generate the initial segmentation masks.
\subsubsection{Human Curation}: A human annotator performs a quick review of the generated masks, deleting obviously incorrect segmentation results. This step does not require experienced engineers or meticulous attention to detail, and avoids the time-consuming effort of pixel-level segmentation mask annotations.
\subsubsection{Prompt-Free Finetuning}: The curated, high-quality masks and their corresponding SEM images are then used to finetune the SAM model from its pre-trained weights for a few epochs. There are two distinct design choices in this step. First, prompt-free finetuning enables the model to learn from its prompt-driven, high-quality outputs, thereby improving convergence speed and reducing deployment costs. Second, we finetune from the original pre-trained weights at each iteration, which prevents overfitting and the catastrophic forgetting problem. These two strategies effectively bridge the gap between prompt-based inference and domain-specific fine-tuning, leading to a more robust and faster adaptation of SAM to our specific domain.
\subsubsection{Weight Update}: The finetuned SAM model weight is then used to update the model in the first step for the subsequent bootstrapping loop.

Our Human-in-the-Loop Bootstrapping approach exploits SAM's general segmentation capability to drive this pipeline. By iteratively learning from its own curated outputs, it replaces costly pixel-level annotation with minimal supervision, significantly reduces annotation time, eliminates the need for professional annotators, achieves faster convergence with only a few epochs, and unifying the advantages of prompt-based and finetuneing strategies for SAM domain adaptation.
  
\subsection{Fine Segmentation via Point-Wise Refinement}

In the fine segmentation stage, we propose a novel refinement approach that leverages lithography and SEM physics to achieve pixel-level accuracy without compromising robustness. Our refinement method is comprised of two steps:

\subsubsection{Feature Extraction}
We first extract the normal vector for each point along the coarse segmentation mask. 
    A slice is sampled along the normal vector, with its scan size determined by the SEM magnification and lithography setup. 
    
    Formally, the scan size $S_{\text{scan}}$ in pixels is given by
    \begin{equation}
    S_{\text{scan}} \;=\;
    \left\lceil
    \frac{ \tfrac{k_{1}\lambda}{\mathrm{NA}} + \Delta_{\text{offset}} }
         { R_{\text{px}}(M) }
    \right\rceil,
    \end{equation}
    where $\tfrac{k_{1}\lambda}{\mathrm{NA}}$ represents the Rayleigh resolution limit, 
    $\Delta_{\text{offset}}$ is a unified term that incorporates practical process and imaging deviations 
    (such as focus/exposure variations, OPC and etch biases, resist effects, and SEM-induced broadening), 
    and $R_{\text{px}}(M) = \tfrac{D_{\text{deflection}}/M}{N_{\text{px}}}$ denotes the physical pixel size under SEM magnification $M$. 
    The ceiling operator ensures that $S_{\text{scan}}$ is an integer number of pixels. 

    The resulting 1D profile is then aligned by centering on the nearest brightest pixel, ensuring accurate error compensation for the coarse-stage segmentation. These 1D features exhibit translational invariance, meaning trenches share similar 1D profiles when aligned along the same normal direction, making them easier to model and train on than the original 2D representation.
    
\subsubsection{Contour Refinement}

Despite their translational invariance, these 1D features remain orientation-dependent due to SEM physics. As shown in Fig.~\ref{fig:mainFig}, vertical lines exhibit weaker contrast because they align with the SEM scan direction. To overcome this, we employ a MLP that predicts the precise positional displacement of each contour point from the extracted features. By harnessing the robustness of supervised learning, this refinement strategy delivers higher accuracy and greater resilience to variability.

Our Point-Wise Refinement method reformulates the 2D segmentation problem into a 1D regression task, thereby reducing the cost of computation, data collection, and annotation. By predicting bounded displacements along predetermined normals, the method inherently enforces geometric consistency, preventing obscure segmentation artifacts such as holes or spurious connections—defects that critically impact the electronic performance of circuits. In this way, our approach enhances the robustness of recent data-driven methods while preserving the metrology-grade precision characteristic of traditional threshold-based techniques.

\section{Experiment}

\subsection{Experiment Setup}
\label{sec:exp_design}

\begin{table}
    \centering
    \caption{Testset partitioning by process familiarity, pattern familiarity, and process window optimality. }
    \begin{tabular}{ccccc}
        \toprule
        Difficulty & Images & Seen Process  &Seen Pattern & Process Window\\
        \midrule
        Easy & 197 & \checkmark  &$\times$& Optimal\\
        Medium & 589 &   $\times$&\checkmark & Optimal\\
        Hard & 197 &   $\times$&$\times$& Optimal\\
        Extreme & 65 &   $\times$&$\times$&Suboptimal\\
        \bottomrule
    \end{tabular}

    \label{tab:dataset_partition}
\end{table}

\begin{figure*}[t]
    \centering
    \includegraphics[width=0.9\linewidth]{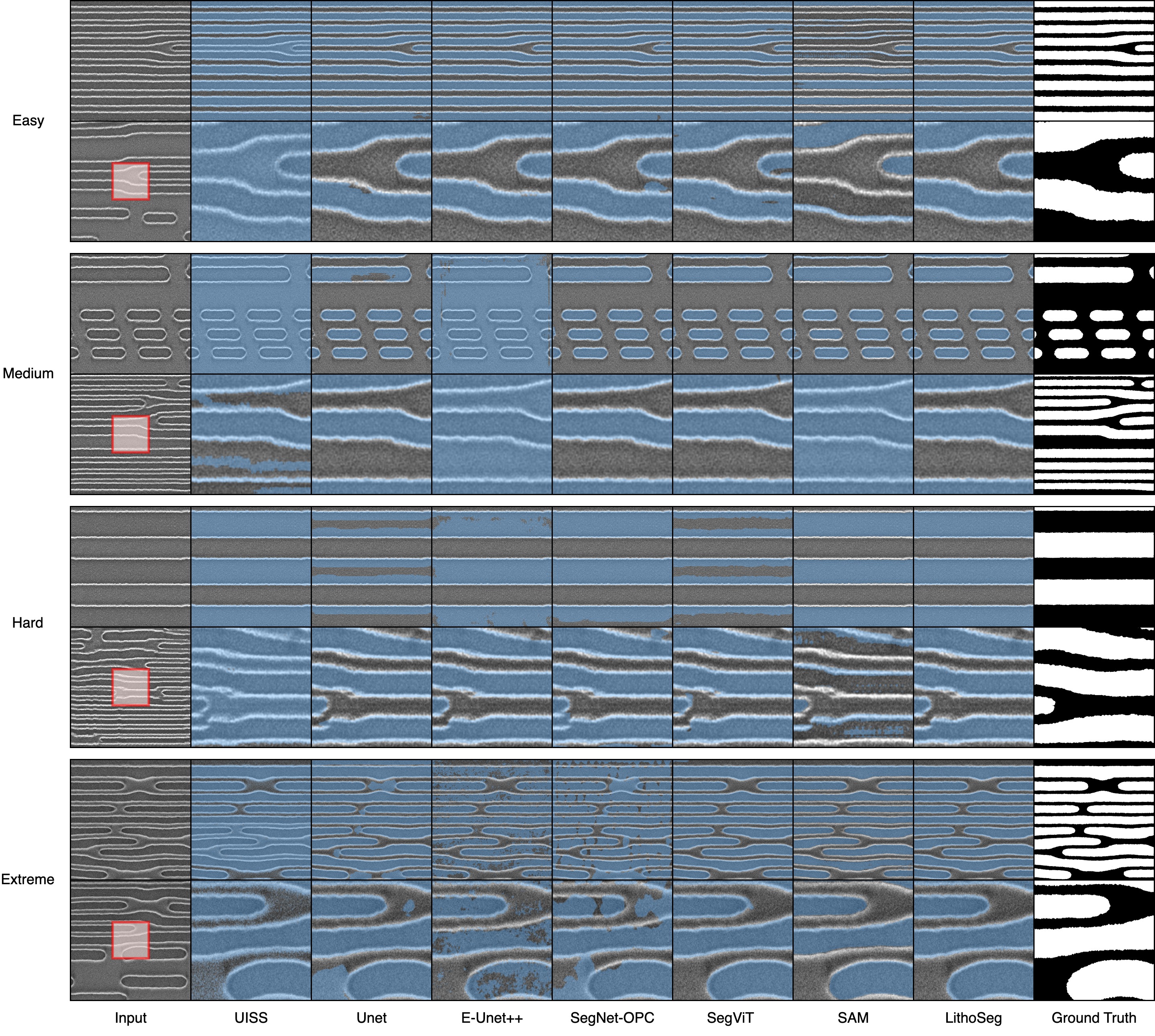}
    \vspace{-10pt} 
    \caption{Comparison of segmentation model performance across different difficulty levels. The columns show five levels of dataset difficulty, from "Easy" to "Extreme". The rows display the original SEM images (input), segmentation output masks overlay on the SEM images, and the ground truth masks. Please zoom in to see the detailed quality differences.}
    \label{fig:visual_compare}
\end{figure*}

\subsubsection{Dataset}

Our dataset includes paired images of circuit layouts, SEM scans and segmentation masks from production lines at $1024 \times 1024$ resolution. Our training dataset consists of 596 images with optimal process window. To evaluate the model's generalization capability, we partitioned the test set into four subsets based on their similarity with the training dataset, as shown in Table~\ref{tab:dataset_partition}. For supervised methods, training stops when the validation loss stops decreasing for five consecutive epochs.

\subsubsection{Implementation Details}
Our model was trained and evaluated on an Ubuntu workstation with i9-14900k and 4090. We used the AdamW optimizer and a combined Dice Loss and Cross-Entropy Loss function for all models. The code and data will be made publicly available on GitHub upon acceptance of the paper. We apply thresholding to extract bounding boxes from layout images. During bootstrapping, the number of training epochs is set to 3 and $S_{scan}$ is set to 30.
 
\subsubsection{Baselines Details}

We compared our LithoSeg method with a wide range of state-of-the-art baselines to cover different model architectures and sizes. Our comparison includes three categories of models: traditional unsupervised method UISS~\cite{UISS}, a weakly-supervised approach (SAM~\cite{Kirillov_2023_ICCV_SAM} with bounding box prompts), and a wide range of state-of-the-art supervised models with different architecture (Unet~\cite{ronneberger2015unetconvolutionalnetworksbiomedical}, E-Unet++~\cite{EUnetpp}, SegNet-OPC~\cite{Xu2025SegNetOPC}, SegViT~\cite{SegViT}).

\subsubsection{Metrics} 

We evaluate LithoSeg across three key aspects: segmentation metrics, electrical metrics, and metrology metrics. \textbf{Segmentation Metrics}: We employ Intersection over Union (IoU), Pixel Accuracy (PA), and F1 score to assess segmentation quality.

\textbf{Electrical Metrics}: To quantify precision in electrical metrics, we evaluate absolute measurement errors between the segmented mask and the ground truth on these electrical metrics.
\begin{itemize}
    \item Open-Short Circuit Connectivity (OSCC)~\cite{ESD}: Quantifies connectivity errors, as in under-segmentation (open circuits) and over-segmentation (short circuits).
    \item Electric Significant Differences (ESD)~\cite{ESD}: Detects partial connectivity defects, such as weak links or unintended parallel traces, which compromise signal integrity.
    \item Critical Dimension (CD)~\cite{Bunday2004}: Measures conductive trace widths to identify fabrication defects.
\end{itemize}
\textbf{Metrology Metrics}: We measures the absolute measurement errors on critical lithography metrology metrics: Line Edge Roughness (LER) and Line Width Roughness (LWR), specifically on Range ($R_E, R_W$), Average Roughness ($R_{Ea},R_{Wa}$), and Root-Mean-Square Roughness ($R^2_{Eq}, R^2_{Wq}$)~\cite{CD-SEM_roughness_parameter}.


\begin{table*}[t]
   \centering
   \caption{Comparison of segmentation, electronic, and metrology metrics across difficulty levels against state-of-the-art models.}
   \resizebox{0.9\textwidth}{!}{  
   \begin{NiceTabular}{llcccccccccccc}
    \toprule 
    \multirow{2}{*}{Test Set} &\multirow{2}{*}{Model}  &\multicolumn{3}{c}{Segmentation}&\multicolumn{3}{c}{Electronic}& \multicolumn{3}{c}{Line Edge Roughness}&\multicolumn{3}{c}{Line Width Roughness}\\
    \cmidrule(lr){3-5} \cmidrule(lr){6-8} \cmidrule(lr){9-11} \cmidrule(lr){12-14}
          &  &  IOU&  PA&  F1 & OSCC & ESD  & CD& $R_E$& $R_{Ea}$& $R^2_{Eq}$& $R_W$& $R_{Wa}$&$R^2_{Wq}$\\
    \midrule
 \multirow{7}{*}{Easy}& UISS & 71.28 & 77.08 & 80.28 & 237.13 & 129.52 & 8.27 & 5.20 & 0.62 & 0.39 & 3.89 & 0.50 & 3.68 \\
  & Unet & 97.36 & 98.54 & 98.23 & 1.67 & 1.57 & -0.05 & -1.05 & -0.05 & -0.03 & -0.70 & -0.04 & -0.18 \\
  & E\mbox{-}Unet++  & 97.63 & 98.69 & 98.42 & 6.35 & 4.94 & 0.05 & -1.35 & -0.10 & -0.06 & -0.87 & -0.08 & -0.35 \\
  & SegNet\mbox{-}OPC  & 97.92 & 98.88 & 98.73 & 0.59 & 0.54 & 0.13 & -1.50 & -0.10 & -0.07 & -1.00 & -0.08 & -0.37 \\
  & SegViT  & 96.23 & 98.06 & 97.65 & 0.58 & 0.17 & -0.50 & -5.25 & -0.47 & -0.30 & -3.43 & -0.33 & -1.37 \\
  & SAM  & 68.93 & 83.78 & 80.77 & 37.52 & 34.68 & -18.55 & -5.35 & -0.63 & -0.40 & -3.28 & -0.40 & -1.53 \\
  & \textbf{LithoSeg} & \textbf{98.13} & \textbf{98.94} & \textbf{98.91} & \textbf{0.25} & \textbf{0.28} & \textbf{-0.12} & \textbf{0.34} & \textbf{0.02} & \textbf{0.01} & \textbf{0.48} & \textbf{0.02} & \textbf{0.31} \\
\hline
 \multirow{7}{*}{Medium}& UISS & 65.41 & 73.71 & 77.37 & 24.01 & 29.41 & 10.22 & 3.14 & 0.51 & 0.33 & 3.39 & 0.47 & 3.12 \\
  & Unet & 91.44 & 95.90 & 95.01 & 1.13 & 1.06 & 0.91 & -1.59 & -0.08 & -0.06 & -0.84 & -0.04 & -0.12 \\
  & E\mbox{-}Unet++  & 43.43 & 45.53 & 59.66 & 80.97 & 99.95 & 16.40 & -3.39 & -0.43 & -0.25 & 1.85 & -0.01 & 2.82 \\
  & SegNet\mbox{-}OPC  & 92.15 & 96.18 & 95.60 & 1.43 & 1.20 & 1.18 & -0.95 & 0.01 & 0.00 & -0.40 & 0.02 & 0.09 \\
  & SegViT  & 91.60 & 95.98 & 95.11 & 0.64 & 0.46 & 0.28 & -5.57 & -0.47 & -0.31 & -3.29 & -0.28 & -0.57 \\
  & SAM  & 70.87 & 85.90 & 82.58 & 25.49 & 21.70 & -19.60 & -6.13 & -0.71 & -0.46 & -3.82 & -0.46 & -1.80 \\
  & \textbf{LithoSeg} & \textbf{93.24} & \textbf{96.77} & \textbf{96.45} & \textbf{0.20} & \textbf{0.13} & \textbf{1.05} & \textbf{-0.89} & \textbf{-0.04} & \textbf{-0.03} & \textbf{-0.35} & \textbf{-0.01} & \textbf{-0.07} \\
\hline
 \multirow{7}{*}{Hard}& UISS  & 69.16 & 75.77 & 79.34 & 223.24 & 84.82 & 10.89 & 1.24 & 0.16 & 0.11 & 0.57 & 0.00 & 0.30 \\
  & Unet  & 89.70 & 94.55 & 93.74 & 1.38 & 1.30 & 3.05 & -0.65 & 0.01 & -0.01 & -2.47 & -0.42 & -1.73 \\
  & E\mbox{-}Unet++  & 89.75 & 94.44 & 93.76 & 15.02 & 6.30 & 3.16 & -0.73 & -0.04 & -0.03 & -2.48 & -0.45 & -1.75 \\
  & SegNet\mbox{-}OPC  & 89.76 & 94.32 & 93.86 & 1.27 & 1.14 & 3.26 & -0.96 & -0.01 & -0.01 & -2.74 & -0.46 & -1.85 \\
  & SegViT  & 89.79 & 94.62 & 93.77 & 2.13 & 1.01 & 2.43 & -5.13 & -0.51 & -0.32 & -5.02 & -0.76 & -2.39 \\
  & SAM  & 62.32 & 80.16 & 75.44 & 57.85 & 55.18 & -17.20 & -4.93 & -0.55 & -0.34 & -4.23 & -0.60 & -2.13 \\
  & \textbf{LithoSeg} & \textbf{90.45} & \textbf{94.84} & \textbf{94.48} & \textbf{0.28} & \textbf{0.76} & \textbf{3.24} & \textbf{0.46} & \textbf{0.03} & \textbf{0.02} & \textbf{-1.66} & \textbf{-0.37} & \textbf{-1.57} \\
\hline
 \multirow{7}{*}{Extreme} & UISS  & 66.39 & 75.64 & 77.21 & 11546.57 & 2480.26 & 11.83 & 3.96 & 0.92 & 0.56 & 5.00 & 0.63 & 3.51 \\
  & Unet  & 70.53 & 74.31 & 79.19 & 15.32 & 25.38 & 0.41 & -1.05 & 0.02 & 0.02 & -0.87 & -0.08 & -0.38 \\
  & E\mbox{-}Unet++  & 62.21 & 76.82 & 75.21 & 2980.54 & 1403.38 & -7.99 & -1.11 & 0.87 & 0.48 & 2.22 & 0.93 & 5.11 \\
  & SegNet\mbox{-}OPC  & 53.26 & 64.27 & 66.28 & 39.23 & 53.43 & 1.02 & 38.28 & 6.59 & 4.40 & 40.28 & 8.37 & 296.54 \\
  & SegViT  & 85.38 & 92.61 & 91.20 & 7.82 & 3.75 & -0.66 & -5.30 & -0.40 & -0.27 & -3.15 & -0.28 & -1.54 \\
  & SAM  & 77.60 & 86.51 & 86.79 & 28.95 & 26.46 & -14.21 & -3.05 & -0.29 & -0.19 & 0.91 & 0.45 & 1.14 \\
  & \textbf{LithoSeg} & \textbf{89.63} & \textbf{90.26} & \textbf{92.28} & \textbf{0.25} & \textbf{0.48} & \textbf{-2.12} & \textbf{0.54} & \textbf{0.07} & \textbf{0.03} & \textbf{0.48} & \textbf{0.05} & \textbf{1.01} \\
  \bottomrule
  \end{NiceTabular}
 }
   \label{tab:experiment_results}
\end{table*}

\subsection{Visual Comparison}

As shown in the Fig.~\ref{fig:visual_compare}, our method significantly outperforms previous approaches, even with supervised methods, while not requiring the segmentation annotation data of the original image. Self-supervised method, UISS, perform worse compare to other approaches. Supervised methods often produces masks with under- or over- segmentation (E-Unet++) and smoothened contours (SegViT), which greatly compromised electrical metrology precision. Our LithoSeg method produces high-quality segmentation masks, demonstrating superior segmentation capability across all difficulty level.

\subsection{Quantitative Comparison}

Table~\ref{tab:experiment_results} presents the quantitative results for segmentation accuracy and absolute measurement errors for both electronic and lithography metrics. Across all levels of difficulty, LithoSeg consistently outperforms all baseline methods. Unsupervised and weakly supervised approaches exhibit stable but modest performance across different difficulty levels. In contrast, fully supervised models achieve strong results on the Easy test set but experience a sharp performance decline as difficulty increases, especially under sub-optimal lithography conditions. Our method demonstrate superior segmentation accuracy and exceptional fidelity in electronic and lithographic metrology across difficulty levels, showing great potential for real-world deployment.

\subsection{Training Efficiency Comparison}

\begin{table}[t]
\centering
\caption{Comparison of Fine-tuning Parameters and Training Time Across Different Models. The proposed LithoSeg model achieves high efficiency, completing its training stages in a significantly shorter total time compared to the baselines.}
\begin{tabular}{lcccc}
\toprule
  Model&Stage& Images & Epochs & Training time \\
\midrule
   SegNet-OPC&-& 596& 28& 15m 24s\\
   E-Unet++&-& 596& 31& 24m 17s\\
   Unet&-& 596& 73& 49m 53s\\
   SegViT&-& 596& 52& 1h 37m\\
 UISS& -& 596& 100& 9h 3m\\
 \hline \hline
  \multirow{5}{*}{\textbf{LithoSeg}}&Coarse 1& 250  &3 & 2m 25s\\
  &Coarse 2& 580  &3 & 5m 32s\\
  &Coarse 3& 596  &3 & 5m 39s\\
   &Fine& 300& 50& 1.94s\\
   \cline{2-5}
 & \textbf{Total}& -&-&\textbf{13m 18s}\\
 \bottomrule
\end{tabular}
\label{tab:efficiency}
\end{table}

\begin{table*}[h]
    \centering
    \caption{Ablation study of the fine-stage refinement module, evaluating the impact of key design choices on the easy test-set. We analyze feature alignment, alternative regressor architectures, robustness to angular perturbations, and feature scan size against the baseline model.}
    \label{tab:combined_results_nicematrix}
    \resizebox{0.9\textwidth}{!}{

    \vspace{1cm}
    \begin{NiceTabular}{lccccccccccccc}
        \toprule
        \Block{2-1}{Experiment} & \Block{2-1}{Params (Mb)} & \Block{1-3}{Segmentation} &&& \Block{1-3}{Electronic} &&& \Block{1-3}{Line Edge Roughness} &&& \Block{1-3}{Line Width Roughness} \\
        \cmidrule(lr){3-5} \cmidrule(lr){6-8} \cmidrule(lr){9-11} \cmidrule(lr){12-14}
        & & IoU & PA & F1 & OSCC & ESD & CD & $R_E$ & $R_{Ea}$ & $R_{Eq}^2$ & $R_W$ & $R_{Wa}$ & $R_{Wq}^2$ \\
        \midrule
        no brightest center        & 0.41 & 77.83 & 89.48 & 87.34 & 0.18 & 0.04 & -16.87 &  2.66 & 0.27 & 0.15 &  2.57 & 0.28 & 1.60 \\
        \hline
        regressor cnn              & 4.54 & 75.37 & 86.97 & 85.80 & 0.17 & 0.42 &   0.19 & -3.00 & -0.28 & -0.18 & -1.23 & -0.16 & -0.49 \\
        regressor transformer      & 2.38 & 75.45 & 87.02 & 85.85 & 0.17 & 0.37 &   0.18 & -3.58 & -0.31 & -0.20 & -1.62 & -0.19 & -0.66 \\
        \hline
        angle perturbation $2^\circ$  & 0.41 & 97.90 & 98.82 & 98.79 & 0.14 & 0.03 &  -0.24 & -0.59 & -0.09 & -0.05 & -0.36 & -0.06 & -0.24 \\
        angle perturbation $5^\circ$  & 0.41 & 97.99 & 98.92 & 98.85 & 0.13 & 0.03 &  -0.24 & -0.61 & -0.09 & -0.05 & -0.41 & -0.06 & -0.24 \\
        angle perturbation $10^\circ$ & 0.41 & 97.78 & 98.76 & 98.73 & 0.13 & 0.02 &  -0.24 & -0.73 & -0.10 & -0.06 & -0.44 & -0.07 & -0.28 \\
        \hline
        scan size 20px             & 0.39 & 81.26 & 90.91 & 89.44 & 0.19 & 0.06 & -12.35 &  9.76 & 0.89 & 0.71 &  9.30 & 0.90 & 9.07 \\
        scan size 40px             & 0.43 & 68.35 & 82.02 & 81.01 & 0.20 & 4.12 &   5.80 &  3.96 & 0.56 & 0.37 &  3.88 & 0.47 & 3.37 \\
        scan size 60px             & 0.47 & 50.81 & 70.04 & 67.14 & 0.26 & 5.75 &   0.65 &  2.13 & 0.22 & 0.17 &  2.27 & 0.26 & 3.71 \\
        \hline\hline
        \textbf{ours}              & 0.41 & \textbf{98.13} & \textbf{98.94} & \textbf{98.91} & \textbf{0.25} & \textbf{0.28} & \textbf{-0.12} & \textbf{0.34} & \textbf{0.02} & \textbf{0.01} & \textbf{0.48} & \textbf{0.02} & \textbf{0.31} \\
        \bottomrule
    \end{NiceTabular}
    }
\end{table*}

In addition to its superior performance, the proposed LithoSeg model offers significant advantages in training efficiency. This is a critical factor when adopt to new manufacturing nodes and is essential for practical applications. Table~\ref{tab:efficiency} summarizes the training times for LithoSeg and several baseline models. LithoSeg's training process is divided into multiple stages. During the coarse stage, the number of high-quality masks increases rapidly as the model converges. In the fine stage, the MLP model's simplicity makes its training time negligible. In total, LithoSeg achieves a total training time of 13 minutes and 18 seconds, nearly 20\% faster than the fastest supervised models. 

Beyond the improvement on computational efficiency, LithoSeg also dramatically reduces the cost in the annotation process by replacing time-consuming, pixel-level annotation with an easy and fast human selection process. In the fine stage, 1D annotation is significant easier to collect and annotate than 2D segmentation annotation. This combined efficiency makes LithoSeg a highly practical solution for real-world lithography metrology, where a fast development cycle is essential.

\subsection{Ablation study}

\subsubsection{Ablation Study on Noise Robustness in the Coarse Stage}

\begin{figure}
    \centering
    \includegraphics[width=0.9\linewidth]{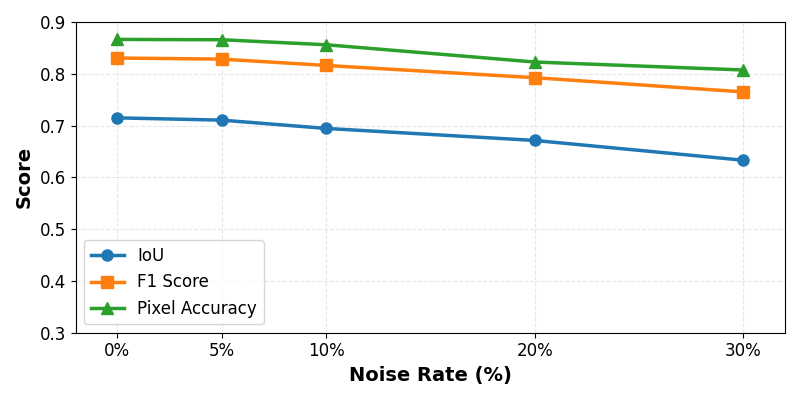}
    \caption{Performance vs. Noise Rate in Bootstrapping Ablation Study}
    \label{fig:bootstrapping}
\end{figure}

In the noise robustness experiment, we evaluated LithoSeg under varying levels of corrupted supervision by injecting 0\%–30\% random noise into the coarse 1 stage training masks and measuring its impact on IoU, F1 Score, and Pixel Accuracy. As shown in the Fig.~\ref{fig:bootstrapping}, performance remains highly stable even under significant noise levels and it shows the proposed Human-in-the-Loop Bootstrapping strategy allows noisy mask inputs from the human curation. This enables vision foundation model, SAM, to be adopted to lithography domain with minimal supervision cost, and enables real-world manufacturing deployment where annotation errors are inevitable.

\subsubsection{Fine stage implementation}

To validate the key design choices in our fine-stage refinement module, we conducted an extensive ablation study, with results presented in Figure \ref{tab:combined_results_nicematrix}. First, removing the brightest-center alignment for feature extraction leads to a substantial drop in segmentation accuracy and large metrology errors, underscoring its importance for small model's regression capability. Second, alternative regressor models proved less effective than our lightweight MLP, despite their higher parameter counts, confirming the MLP’s superior efficiency for this 1D positional refinement task. The model is robust to upstream errors, showing only graceful degradation with angular perturbations up to 10$^\circ$. Finally, the scan size of the 1D feature proved critical, with any deviation from the optimal length severely degrading accuracy across all metrics. Collectively, these findings validate that our fine-stage design, which combines brightest-pixel centering, a lightweight MLP regressor, and a precise scan size derived from SEM magnification and lithography machine, is essential for achieving state-of-the-art precision in lithography segmentation.

\section{RELATED WORKS}

\subsection{Lithography Segmentation}

Automated lithography segmentation and metrology has been a long-stranding challenge in semiconductor manufacturing. Traditional approaches, which use thresholding on 1D features along the groove normals, are valued for their high metrology precision but necessitates manual intervention for point selection every time~\cite{Hector2005, CD-SEM_roughness_parameter}, which limits their use to a tool-based role. In contrast, supervised deep learning methods automate the segmentation process but lack the pixel-level precision required for accurate metrology, as they tend to produce overly smoothed edge contours, limiting their applicability~\cite{UISS, ronneberger2015unetconvolutionalnetworksbiomedical, SegViT, Xu2025SegNetOPC, EUnetpp, Jacob2023, learning_to_detect_defect}. Our work addresses this critical gap by developing a hybrid approach that leverages the strengths of both traditional model-based methods and supervised data-driven approaches, allowing us to automate the segmentation process with metrology-grade precision.

\subsection{Visual Foundation Models}
A major bottleneck for fully-supervised segmentation methods is the high cost of collecting and annotating large datasets. The recent advent of vision foundation models, particularly the Segment Anything Model (SAM)~\cite{Kirillov_2023_ICCV_SAM}, has introduced a new paradigm with its impressive general segmentation capabilities. Transferring SAM to specific domains through prompt-based or finetune-based approaches, has emerged as a low-cost alternative to training networks from scratch~\cite{Ren_2024_WACV, Zhang_Liu_Lin_Liao_Li_2024, zhang2023segmentmodelsamradiation, ali2024evaluating, archit2025segment, ma2024segment}. We build upon these efforts by proposing a novel Human-in-the-Loop Bootstrapping process, which combines prompting and finetuning in an iterative loop, enabling rapid convergence and high performance with minimal human supervision.

\section{Conclusion}

We propose LithoSeg, a novel two-stage, coarse-to-fine framework that provides a scalable and high-precision solution for lithography segmentation and metrology. By leveraging a human-in-the-loop finetuning strategy on  Segment Anything Model (SAM) for robust coarse segmentation and then refining contours with a lightweight MLP network, our method achieves an optimal balance of superior generalization and metrology-grade precision with human supervision. Our experiments confirm that LithoSeg outperforms state-of-the-art baselines and demonstrates exceptional robustness to new patterns and non-optimal process window, making it a highly practical solution for lithography metrology in semiconductor manufacturing.

\vfill\pagebreak



\bibliographystyle{IEEEbib}
\bibliography{references}

\end{document}